\documentclass[pdflatex,sn-mathphys-num]{sn-jnl}

\usepackage{graphicx,tabularx}%
\usepackage{multirow}%
\usepackage{amsmath,amssymb,amsfonts}%
\usepackage{amsthm}%
\usepackage{mathrsfs}%
\usepackage[title]{appendix}%
\usepackage{xcolor}%
\usepackage{textcomp}%
\usepackage{manyfoot}%
\usepackage{booktabs}%
\usepackage{algorithm}%
\usepackage{algorithmicx}%
\usepackage{algpseudocode}%
\usepackage{listings}%

\usepackage{pifont}

\theoremstyle{thmstyleone}%
\theoremstyle{thmstyletwo}%

\theoremstyle{thmstylethree}%

\newcommand\figref{Fig.~\ref}
\newcommand\tabref{Table~\ref}
\newcommand\secref{Section~\ref}
\newcommand\equref{Eq.~\ref}

\begin{document}

\title[Confidence-aware Monocular Depth Estimation]{Confidence-aware Monocular Depth Estimation for Minimally Invasive Surgery}


\author*[1]{\fnm{Muhammad} \sur{Asad}}\email{muhammad.asad@medtronic.com}
\author[1]{\fnm{Emanuele} \sur{Colleoni}}
\author[1]{\fnm{Pritesh} \sur{Mehta}}
\author[1]{\fnm{Nicolas} \sur{Toussaint}}
\author[1]{\fnm{Ricardo} \sur{Sanchez-Matilla}}
\author[1]{\fnm{Maria} \sur{Robu}}
\author[1]{\fnm{Faisal} \sur{Bashir}}
\author[1]{\fnm{Rahim} \sur{Mohammadi}}
\author[1,2]{\fnm{Imanol} \sur{Luengo}}
\author[1,2]{\fnm{Danail} \sur{Stoyanov}}

\affil[1]{\orgname{Medtronic Surgical}, \orgaddress{\city{London}, \country{UK}}}

\affil[2]{\orgdiv{UCL Hawkes Institute}, \orgname{University College London}, \orgaddress{\city{London},\country{UK}}}






\abstract{
\textbf{Purpose:} Monocular depth estimation (MDE) is vital for scene understanding in minimally invasive surgery (MIS). However, endoscopic video sequences are often contaminated by smoke, specular reflections, blur, and occlusions, limiting the accuracy of MDE models. In addition, current MDE models do not output depth confidence, which could be a valuable tool for improving their clinical reliability. 

\textbf{Methods:} We propose a novel confidence-aware MDE framework featuring three significant contributions: (i) Calibrated confidence targets: an ensemble of fine-tuned stereo matching models is used to capture disparity variance into pixel-wise confidence probabilities; (ii) Confidence-aware loss: Baseline MDE models are optimized with confidence-aware loss functions, utilizing pixel-wise confidence probabilities such that reliable pixels dominate training; and (iii) Inference-time confidence: a confidence estimation head is proposed with two convolution layers to predict per-pixel confidence at inference, enabling assessment of depth reliability.

\textbf{Results:} Comprehensive experimental validation across internal and public datasets demonstrates that our framework improves depth estimation accuracy and can robustly quantify the prediction's confidence. On the internal clinical endoscopic dataset (StereoKP), we improve dense depth estimation accuracy by $\approx$8\% as compared to the baseline model. 

\textbf{Conclusion:} Our confidence-aware framework enables improved accuracy of MDE models in MIS, addressing challenges posed by noise and artifacts in pre-clinical and clinical data, and allows MDE models to provide confidence maps that may be used to improve their reliability for clinical applications.
}


\maketitle
\begin{figure}[ht]
    \centering
    \includegraphics[width=1.0\linewidth]{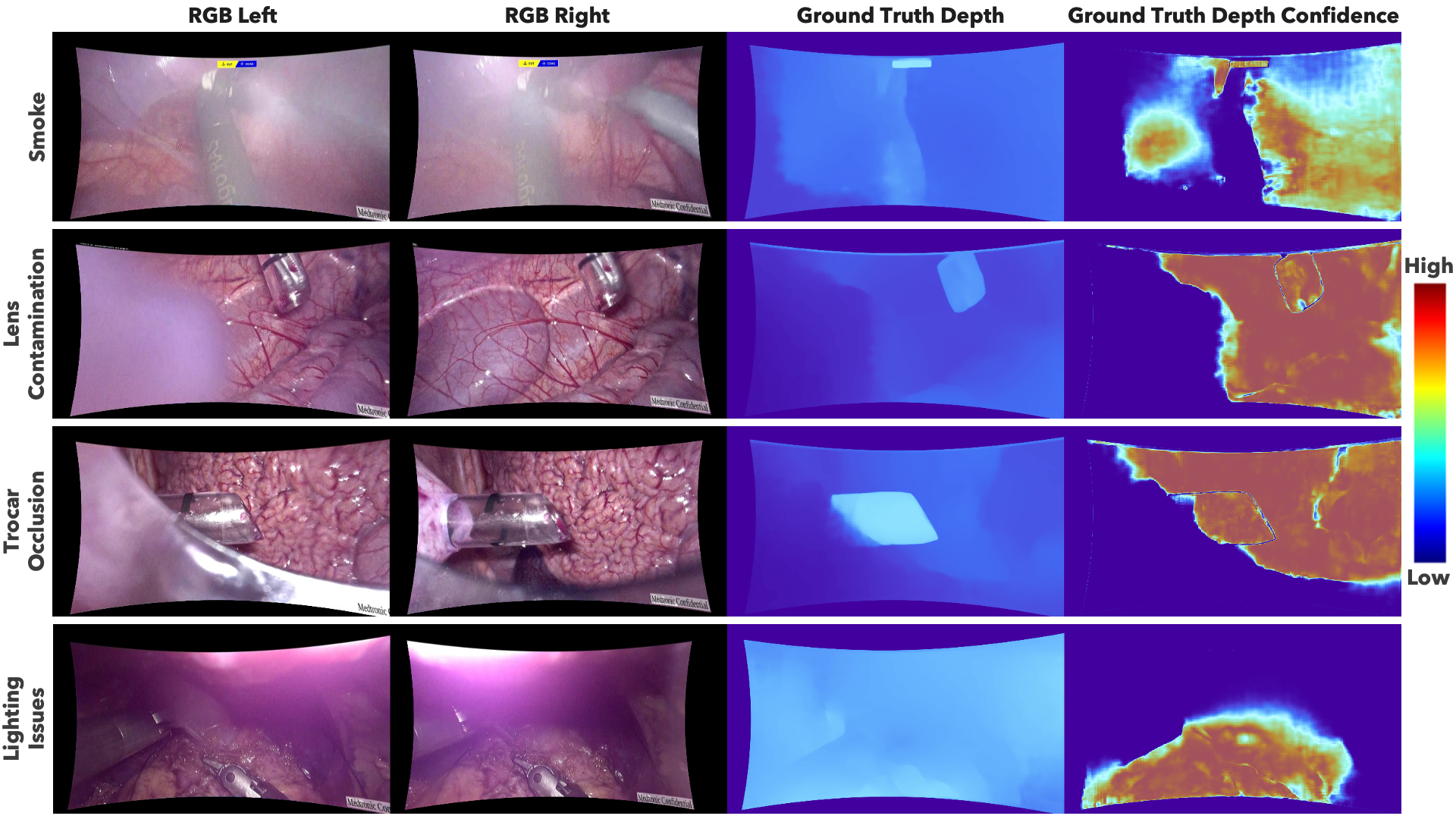}
    \caption{Factors such as image acquisition noise, lens contamination/smudges, blur, occluded views, and lighting issues in MIS datasets contribute towards unreliable depth data that impacts MDE models trained on MIS datasets. Depth confidence can be used to identify noise-free regions in confidence-aware training. Examples from the StereoKP dataset.}
    \label{fig:issues_in_mis_wild_data}
\end{figure}

\section{Introduction}
\label{sec:intro}
Minimally invasive surgery (MIS) increasingly relies on computer vision to perceive 3D scene geometry for tasks such as surgical navigation assistance, autonomous tissue manipulation, safety monitoring, and anatomical measurements \cite{moghani2025sufia,pratt2010dynamic}. Accurate monocular depth estimation (MDE) is particularly compelling in this context, as it is compatible with the monocular endoscope prevalent in clinical practice. However, endoscopic images are especially challenging for depth models because they often contain non-uniform lighting, shiny surfaces (specular highlights), smoke, fluids, occlusions from surgical instruments, and fast camera movements (as shown in \figref{fig:issues_in_mis_wild_data}). These factors disrupt the visual consistency and assumptions, such as uniform reflectance and steady illumination, that many depth estimation models rely on. In surgical settings, knowing when a depth prediction is unreliable is as crucial as achieving low average error. This is because unreliable depth predictions can lead to incorrect anatomical assessments or unsafe instrument navigation, potentially increasing the risk of surgical errors and compromising patient safety. This work addresses both accuracy and reliability by integrating explicit pixel-wise confidence into the training and inference of MDE in the MIS context.

\noindent \textbf{Background and related work:} Classical surgical 3D methods take advantage of calibrated stereo rigs and triangulation or multiview geometry (e.g., SLAM, SfM) to recover dense reconstructions of the patient's anatomy \cite{lou2024ws,shao2022self}. While capable of producing metric depth maps, these pipelines are sensitive to calibration drift, texture scarcity in soft tissue, narrow baselines, and endoscope-specific artifacts; in practice, they often require careful tuning and may underperform in the presence of smoke or heavy specular reflections \cite{budd2024transferring}. Learning-based depth approaches have been recently proposed to advance endoscopic reconstruction, including supervised models trained on synthetic or proxy labels \cite{budd2024transferring,toussaint2025zero}, self-supervised monocular approaches using photometric consistency \cite{ye2017self}, and domain-adaptation techniques that transfer exogenous priors to the surgical domain \cite{mahmood2018unsupervised}. Building on these advances in depth estimation, recent research has also focused on quantifying uncertainty in regression models, which is increasingly relevant for surgical applications where reliable confidence estimates are critical for clinical decision-making. This includes models that predict observation noise \cite{kendall2017uncertainties}, Bayesian approximations via Monte Carlo dropout \cite{gal2016dropout}, deep ensembles \cite{lakshminarayanan2017simple}, and post-hoc calibration \cite{kuleshov2018accurate} to relate predicted confidence to actual error. Budd et al. \cite{budd2024transferring} generate pseudo ground truth from stereo endoscopic video by rectifying frames, applying bidirectional optical flow, and supervising only on pixels meeting a 2-px cycle-consistency check. This binary masking lacks graded reliability, assumes strict rectification and rigidity, and excludes challenging regions, resulting in supervision gaps and selection bias. Xu et al. \cite{xu2024daua} pre-train a depth model on a small labeled set and pseudolabel the rest using zero-shot inference, with uncertainty estimated via test-time augmentation (TTA) dispersion. The per-pixel standard deviation is scaled to [0, 1] and used as loss weights. However, this method lacks an uncertainty head, omits stereo cues for challenging regions, and does not calibrate standard deviation to confidence, allowing large deviations to dominate without probabilistic meaning. In contrast, our approach uses stereo evidence, learns per-pixel confidence, and calibrates it with a scaling parameter.

\noindent \textbf{This work:} We propose a confidence-aware supervised training framework for monocular surgical depth estimation in surgical scenes. We first compute pixel-wise confidence labels by running a diverse ensemble of stereo-matching models on surgical video frames and converting their variance into confidence scores. These confidence labels are then used in conjunction with monocular images and depth supervision to train standard MDE backbones. Losses are weighted by confidence to suppress the influence of unreliable depth supervision and to emphasize trustworthy regions during learning.  In addition, we attach a lightweight confidence head to the MDE backbone and supervise it directly with the ensemble-derived confidence labels, allowing the model to output per-pixel confidence at inference alongside depth. This explicit confidence estimation supports downstream applications, such as surgical navigation assistance, autonomous tissue manipulation, safety monitoring, and anatomical measurements, by providing reliability measures that help mitigate risks associated with uncertain predictions.

\noindent \textbf{Contributions: }
We address gaps in existing MDE methods with the following main contributions of this work:
\begin{itemize}
\item \textbf{Confidence map.} 
We compute pixel-wise depth confidence using an ensemble of stereo matching models and introduce a novel function to convert variance into probability, yielding continuous confidence maps that closely correspond to the actual depth confidence. looseness = -1
\item \textbf{Confidence-aware loss.} We incorporate the calibrated confidence maps into MDE training via confidence-weighted losses, emphasizing reliable regions while down-weighting uncertain or noisy pixels.
\item \textbf{Confidence at inference.} We attach a lightweight confidence head, enabling per-pixel confidence prediction at inference to improve the reliability of MDE in clinical settings.\looseness=-1
\end{itemize}

\begin{figure}[t]
    \centering
    \includegraphics[width=0.7\linewidth]{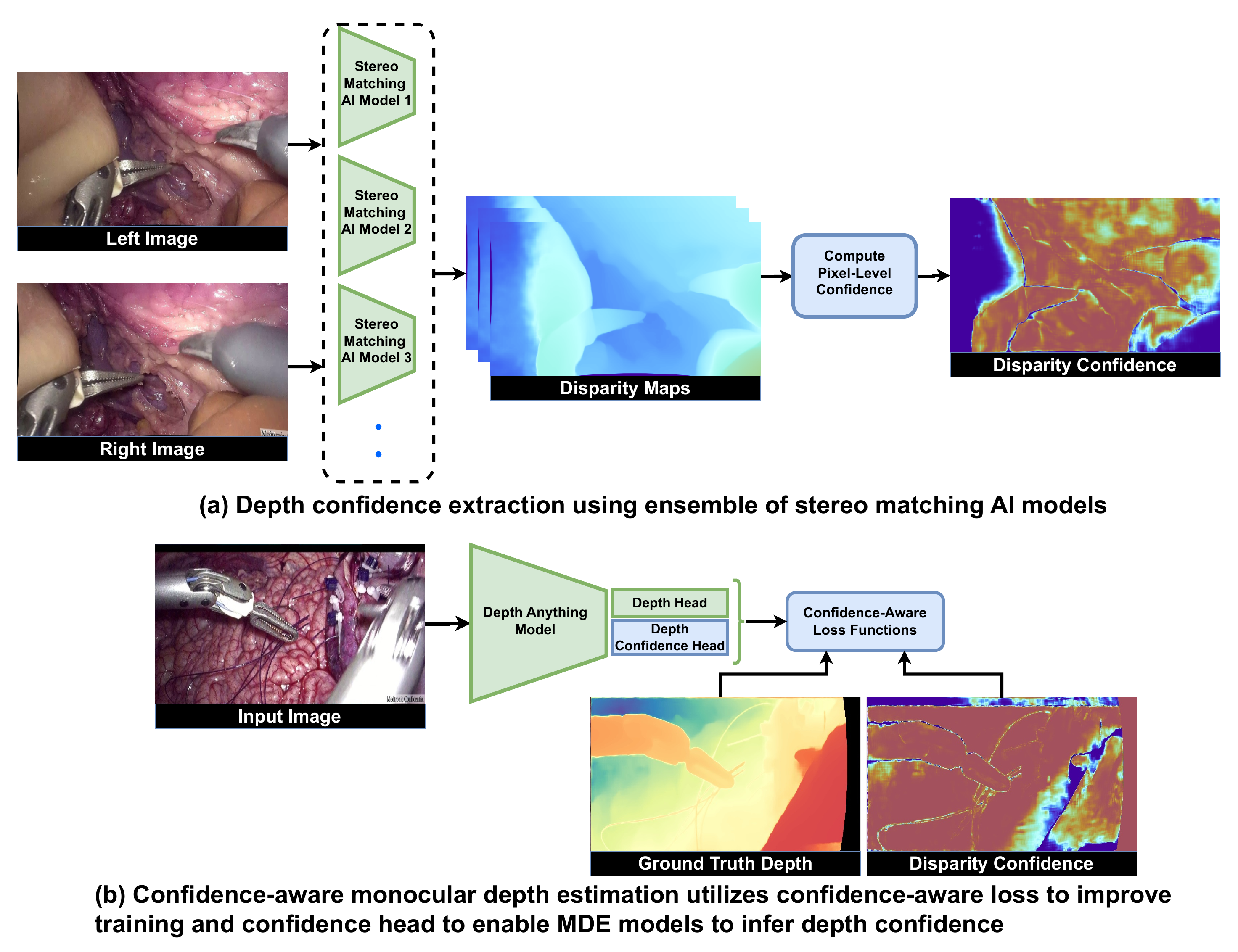}
    \caption{The proposed confidence-aware training methods. Our proposed blocks are shown in blue. Method (a) is used to add depth confidence to a depth dataset, whereas (b) shows our confidence-aware MDE training framework. }
    \label{fig:confidence-aware_overview}
\end{figure}

\section{Methods}
\label{sec:methods}
Our confidence-aware MDE pipeline is shown in Figure 2. First, pairs of stereo frames are input to an ensemble of stereo matching models to produce disparity maps. Next, disparity confidence is calculated by aggregating predictions across the ensemble. Finally, groundtruth depth and the associated disparity confidence are used with the proposed confidence-aware loss to train an MDE model with two heads: one for depth and one for per-pixel confidence. Below, we detail each component.

\subsection{Ensemble-based Depth Confidence Estimation}
\label{sec:ensemble_conf}
Quality degrades in MIS scenes with motion blur, smoke/smudges, exposure extremes, occlusions, or restricted views (\figref{fig:issues_in_mis_wild_data}). To identify image regions with such challenges, we train an ensemble of $K$ stereo matching models \cite{xu2023unifying}, pre-trained on natural images and fine-tuned on MIS stereo data with different random seeds \cite{lakshminarayanan2017simple}. The ensemble yields $K$ disparity maps ${D^{k}}_{k=1}^K$ per frame. We compute the per-pixel variance $D_v = \frac{1}{K} \sum_{k=1}^K \left(D^k - \hat{D} \right)^2$, where $\hat{D}$ is per-pixel mean across $K$ ensembles, and convert it to confidence using:
\begin{equation} 
P_c(i) = \exp\left(-\frac{D_v(i)}{2\sigma^2}\right),
\label{eq:variance_to_confidence_conversion} 
\end{equation} 
where $\sigma$ controls the softness of the confidence mapping, a larger $\sigma$ makes the confidence less sensitive to variance, while a smaller $\sigma$ results in sharper drops in confidence for noisy regions. Intuitively, ambiguous/noisy regions produce higher ensemble variance (lower confidence), whereas artifact-free areas yield lower variance (higher confidence).

\subsection{Confidence-aware Monocular Depth Estimation}
We use the confidence estimated in \secref{sec:ensemble_conf} to train a monocular depth model \cite{yang2024depth} using the proposed confidence-aware loss functions and a confidence-prediction head.

\subsubsection{Confidence-aware Loss}
A confidence-aware loss function is introduced to incorporate uncertainty into MDE models during training, thereby improving their robustness and reliability in challenging, noisy MIS environments, as demonstrated by our experiments. Let $l_i$ be the loss contribution for pixel $i$ (e.g., depth regression error), and $P_c(i)$ the corresponding confidence, as determined by our ensemble-based confidence estimate (see Section~\ref{sec:ensemble_conf}). The general form of the confidence-aware loss is then given by:
\begin{equation}
\mathcal{L}_{\text{conf}} = \frac{1}{N} \sum_{i=1}^N P_c(i) \cdot l_i,
\label{eq:confidence_aware_loss}
\end{equation}
where $N$ is the total number of pixels with depth information. This formulation ensures that regions with higher depth confidence contribute more to training, while regions with noise are down-weighted.
We apply confidence-based weighting (from \equref{eq:confidence_aware_loss}) to three key loss functions as:
\begin{equation}
    \mathcal{L}_{\text{total}} = \mathcal{L}_{\text{silog\_conf}} + \mathcal{L}_{\text{grad\_conf}} + \mathcal{L}_{\text{edge\_conf}},
\end{equation}
where $\mathcal{L}_{\text{silog\_conf}}$ is scale-invariant logarithmic loss \cite{eigen2014depth}, $\mathcal{L}_{\text{grad\_conf}}$ is scale-invariant gradient matching loss \cite{ranftl2020towards}, and $\mathcal{L}_{\text{edge\_conf}}$ is edge-aware smoothness loss \cite{tomasi1998bilateral}.

\subsubsection{Confidence Head Prediction}
As shown in \figref{fig:confidence-aware_overview}(b), we augment the MDE decoder with a confidence head: a $3\times3$ conv (32 channels, ReLU) followed by a $1\times1$ conv producing a single-channel map in $[0,1]$. We train this head with binary cross-entropy \cite{rumelhart1986learning}, enabling inference-time confidence alongside depth.

\section{Experimental Setup}
\begin{table}[t]
\centering
\caption{Summary of stereo endoscopic datasets utilized for experimental validation of proposed confidence-aware MDE methods.}
\begin{tabular}{l p{3cm} p{1.5cm} p{3cm} c}
\hline
\textbf{Dataset} & \textbf{Description} & \textbf{Setting} & \textbf{Modalities} & \textbf{Total Frames} \\
\hline
StereoKP & Internal, 57 pre-clinical + 6 clinical videos & Pre-clinical + Clinical & Metric depth using \cite{xu2023unifying} +  manually annotated instrument keypoints & 10,491 \\
\hline
MicroCT-SE & Internal, surrogate and two ex-vivo liver & Lab & Metric depth from MicroCT & 16,775 \\
\hline
MicroCT-PK & Internal, ex-vivo porcine kidney & Lab & Metric depth using from MicroCT & 4046 \\
\hline
Hamlyn \cite{recasens2021endo} & Public in-vivo stereo endoscope videos \cite{recasens2021endo}  & Clinical & Metric depth using stereo-matching \cite{recasens2021endo} & 92,673 \\
\hline
DaVinci \cite{ye2017self} & Public in-vivo stereo endoscope videos \cite{ye2017self}  & Clinical & Metric depth using \cite{xu2023unifying} & 41,433 \\
\hline
\end{tabular}
\label{tab:dataset_summary}
\end{table}

\subsection{Datasets}
\begin{figure}[t]
    \centering
    \includegraphics[width=0.9\linewidth]{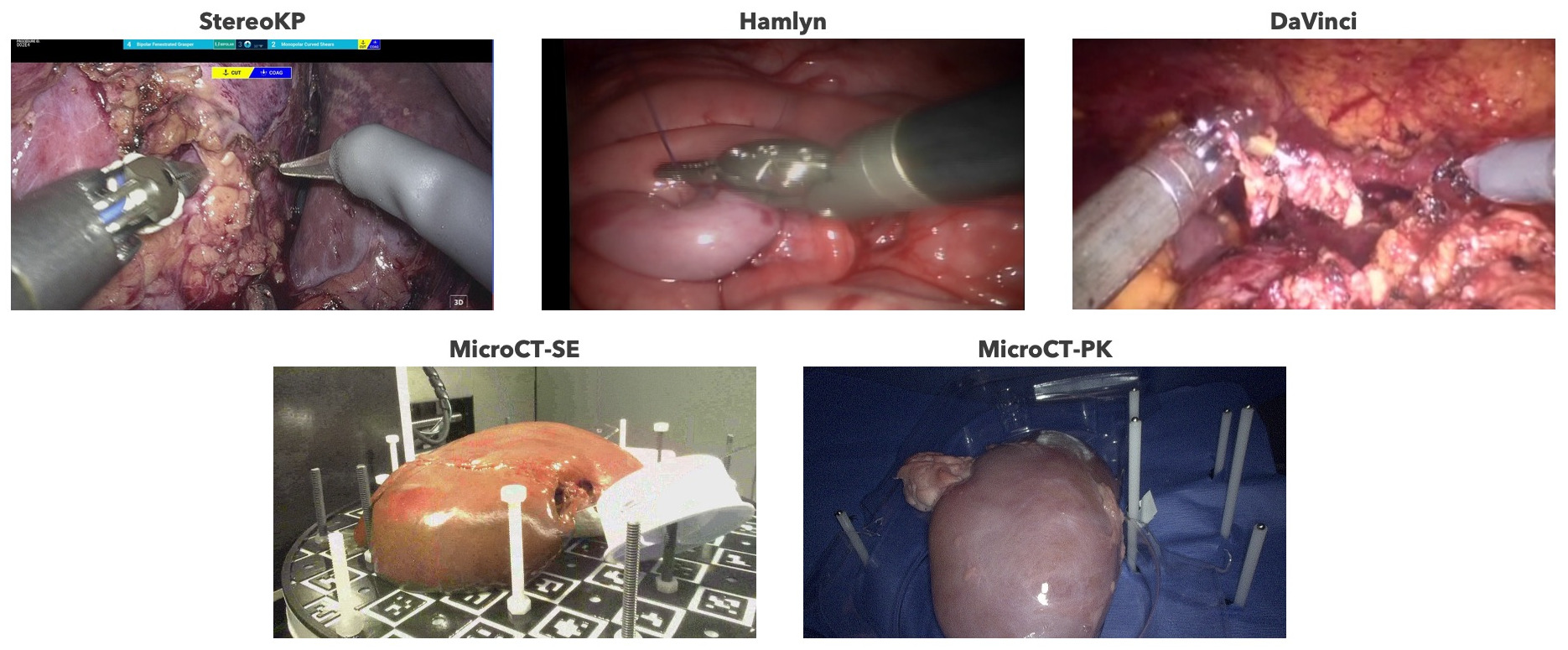}
    \caption{Comparison of frames from different datasets used in this work. StereoKP, Hamlyn \cite{recasens2021endo} and DaVinci \cite{ye2017self} are from clinical settings whereas both MicroCT-SE and MicroCT-PK are from lab settings. StereoKP and MicroCT datasets are internal wheres Hamlyn and DaVinci are public datasets.\looseness =-1}
    \label{fig:compare_images_datasets}
\end{figure}

Datasets used in this work are summarized in \tabref{tab:dataset_summary} and frames compared in \figref{fig:compare_images_datasets}.
For each internal and public dataset, we apply the method in \secref{sec:ensemble_conf} to rectified stereo frames to estimate ensemble-based depth confidence. All datasets provide ground-truth metric depth for training and evaluation. Additionally, StereoKP includes manually annotated 3D keypoints, used only for evaluation. 

\noindent \textbf{StereoKP:}
Internal dataset from the Hugo\textsuperscript{TM} RAS system \cite{ngu2024narrative} with 10,491 stereo frames from 57 pre-clinical and 6 clinical videos. We use 52 pre-clinical and 5 clinical videos for training, 5 pre-clinical for validation, and report results on one randomly selected unseen clinical test video. Dense metric depth is obtained via Unimatch stereo matching \cite{xu2023unifying} with camera parameters. Hugo surgical instrument \cite{ngu2024narrative} keypoints are annotated in left/right frames and triangulated for precise 3D locations, serving as an additional reliable evaluation source.

\noindent \textbf{MicroCT:}
Internal dataset with MicroCT-derived metric depth. (i) \textbf{MicroCT-SE}: 16,775 monocular endoscopic images with ground-truth depth for a surrogate and two ex-vivo livers. (ii) \textbf{MicroCT-PK}: 4,046 monocular images with ground-truth depth of an ex-vivo porcine kidney. For MicroCT-SE, one organ is used for validation and the rest for training; for MicroCT-PK, one sequence is held out for validation and the remaining for training.

\noindent \textbf{Hamlyn \cite{recasens2021endo}:}
Public in-vivo rectified stereo dataset with 21 videos (92,673 frames) exhibiting weak textures, deformations, reflections, tools, and occlusions. Following \cite{recasens2021endo}, videos 1, 4, 19, and 20 are used for validation; the remaining 17 for training.

\noindent \textbf{DaVinci \cite{ye2017self}:}
Public in-vivo rectified stereo dataset with 41,433 image pairs and camera parameters in 11 sequences. We used 7,192 frames from unseen videos for testing and the remainder for training as in \cite{ye2017self}. Since no depth is provided, groundtruth depth is computed using Unimatch \cite{xu2023unifying} with the given camera parameters.

\subsection{Metrics}
We evaluate relative depth predictions using two standard scale-invariant metrics: absolute relative error (ARE) and $\delta_1$ \cite{yang2024depth,toussaint2025zero}. ARE ($\text{ARE} = |d - d^*| / d^*$) measures the average proportional difference between predicted ($d$) and ground-truth depth ($d^*$), while $\delta_1$ computes the percentage of pixels where $\max(d / d^*, d^* / d) < 1.25$, representing the fraction of sufficiently accurate predictions. Following~\cite{toussaint2025zero}, predicted depth maps are median-scaled to correct for scale ambiguity.

Metric depth's accuracy is further assessed using mean absolute error (MAE, in mm) and accuracy within 2\,mm (Acc@2mm), defined as the percentage of samples with depth error $\leq 2$\,mm. These metrics are applied only to the StereoKP dataset, where manually annotated 3D instrument keypoints provide a reliable and independent ground-truth reference.

\subsection{Stereo-Matching Ensemble Fine-tuning}
We fine-tuned five Unimatch models \cite{xu2023unifying}, pretrained on natural images, using a subset of pre-clinical videos from the StereoKP dataset comprising 6,843 frames and 15,000 synthetic stereo pairs generated using the pipeline proposed in 
~\cite{toussaint2025zero} for the Hugo\textsuperscript{TM} RAS system \cite{ngu2024narrative}. Each model was fine-tuned with a unique random seed for 10 epochs using a learning rate of $1{\times}10^{-7}$. The ensemble outputs were combined and processed using Eq.~(\ref{eq:variance_to_confidence_conversion}) to compute per-pixel depth confidence across all datasets.

\subsection{MDE Model Training}
We trained two models in all experiments: (i) the baseline DepthAnything v1–Base (DAv1-B) \cite{yang2024depth} and (ii) a confidence-aware variant (DAv1-B-CA) incorporating the proposed loss and confidence head. Both models were trained for 90 epochs with a learning rate of $1{\times}10^{-5}$ using the OneCycleLR scheduler \cite{smith2019super}, which linearly increases the learning rate during the first 1\% of steps and gradually anneals it thereafter to improve convergence stability.

\section{Results}
\label{sec:results}

\subsection{Hyperparameter and Ablation Study}
For both studies presented in the following sections, we utilize the StereoKP dataset to evaluate and compare each variant across varying settings listed in each experiment.

\begin{table}[ht]
\centering
\caption{Sensitivity analysis of the hyperparameter $\sigma$ in the confidence probability equation (Eq. \ref{eq:variance_to_confidence_conversion}), showing the effect on ARE and $\delta_1 < 1.25$ metrics using the StereoKP dataset.}
\begin{tabular}{l @{\hspace{3.5em}} c @{\hspace{3.5em}} c}
\hline
$\boldsymbol{\sigma}$ & \textbf{ARE (\%) $\downarrow$} & $\boldsymbol{\delta_1 < 1.25}$ (\%) $\uparrow$ \\
\hline
0.2 & 9.58 & 92.68 \\
0.5 & 8.75 & 94.00 \\
0.7 & \textbf{8.86} & \textbf{94.14} \\
1.0 & 9.04 & 93.40 \\
\hline
\end{tabular}
\label{tab:hyperparameter_tuning}
\end{table}

\noindent \textbf{Hyperparameter Study:}
A sensitivity analysis of the parameter $\sigma$ in \equref{eq:variance_to_confidence_conversion} (\tabref{tab:hyperparameter_tuning}) shows that ARE is minimized at $\sigma = 0.5$ and $\delta_1 < 1.25$ peaks at $\sigma = 0.7$, confirming that $\sigma$ strongly affects confidence estimation and model precision. We used $\sigma = 0.7$ in all subsequent experiments, scaled with image resolution to ensure consistent confidence estimation across datasets.

\noindent \textbf{Ablation Study:}
We analyzed the effect of the confidence-aware loss (CAL) and confidence head (CH) across four configurations (\tabref{tab:ablation_components}). Both components individually improved all metrics, with their combination yielding the best overall results, demonstrating complementary benefits where CAL enhances uncertainty handling, while CH improves per-pixel depth reliability.

\begin{table}[ht]
\centering
\caption{Ablation study showing the impact of Confidence Head (CH) and Confidence-aware Loss (CAL) on model performance across four configurations. Results demonstrate that both components, individually and combined, improve MAE, accuracy at 2mm, ARE, and $\delta_1 < 1.25$ over the baseline. Best results are shown in \textbf{bold}.}
\begin{tabular}{c c c c c c}
\hline
\textbf{CH} & \textbf{CAL} & \textbf{MAE (mm) $\downarrow$} & \textbf{Acc @ 2mm (\%) $\uparrow$} & \textbf{ARE (\%) $\downarrow$} & $\boldsymbol{\delta_1 < 1.25}$ (\%) $\uparrow$ \\
\hline
\ding{55} & \ding{55} & 2.04 & 72.4 & 12.41 & 85.83 \\
\ding{55} & \ding{51} & \textbf{1.76} & \textbf{78.2} & 8.99 & 93.62 \\
\ding{51} & \ding{55} & 1.81 & 76.5 & 9.20 & 93.63 \\
\ding{51} & \ding{51} & 1.79 & 77.9 & \textbf{8.86} & \textbf{94.14} \\
\hline
\end{tabular}
\label{tab:ablation_components}
\end{table}

\subsection{Experimental Comparison}
\tabref{tab:main_experimental_comparison} presents a detailed comparison of dense ARE and $\delta_1$ for the baseline DepthAnything V1–Base model \cite{yang2024depth} (DAv1-B) and the proposed confidence-aware variant (DAv1-B-CA) across multiple datasets. 

For the MicroCT-SE and MicroCT-PK datasets, the predicted depths closely match the gold-standard measurements from MicroCT, showing slight but consistent improvements over the baseline. Since these datasets were collected under controlled laboratory conditions with minimal noise, large performance gains are not expected. Nonetheless, the strong agreement with precise MicroCT-derived ground truth confirms the high accuracy and reliability of our MDE models in well-calibrated experimental settings.

For the Hamlyn dataset, improvements are less pronounced than in StereoKP, primarily because its depth maps were preprocessed to remove noise and artifact-prone regions \cite{recasens2021endo}. Since our confidence-aware method primarily benefits ambiguous or uncertain regions, its impact is reduced on manually cleaned data.

In the DaVinci dataset \cite{ye2017self}, we observe consistent but moderate gains, reflecting the inherently higher quality and stability of images captured under well-controlled surgical conditions. Unlike StereoKP, which includes unfiltered, in-the-wild recordings, DaVinci data exhibit minimal illumination variation and texture noise. The observed improvements nonetheless indicate that our model enhances depth precision even in structured, low-noise clinical scenarios, confirming its adaptability across diverse MIS environments.

\begin{figure}[ht]
    \centering
    \includegraphics[width=1.0\linewidth]{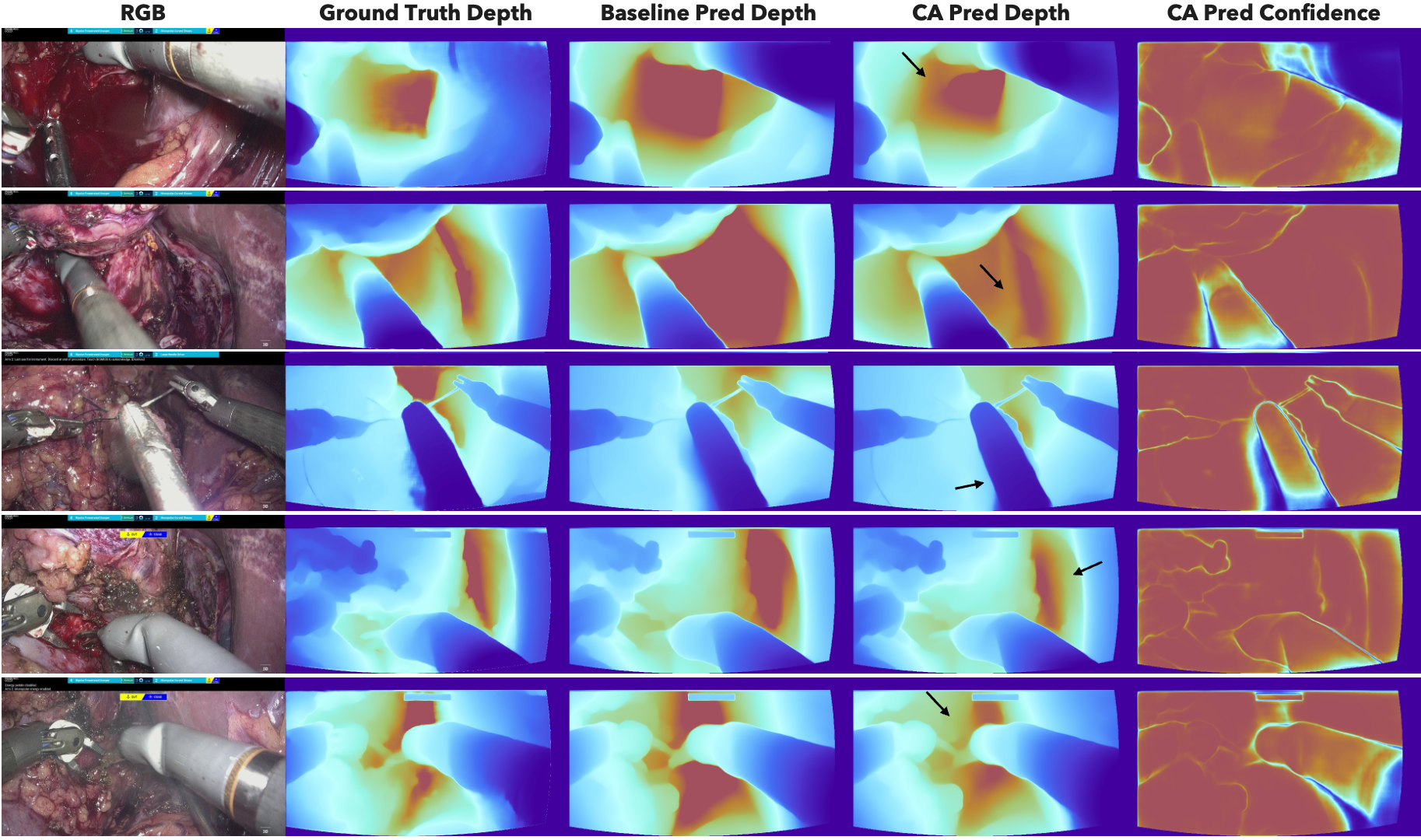}
    \caption{Qualitative results on StereoKP dataset, shows ground truth and predicted depth from DAv1-B baseline and DAv1-B CA (proposed) models, additionally showing predicted confidence from CA model. Arrows show regions where CA model performed particularly better as compared to baseline.}
    \label{fig:qualitative_comparison}
\end{figure}

In the StereoKP dataset, the proposed method shows notable improvements in both keypoint-based and dense metrics, reducing MAE from 2.04mm to 1.79mm and increasing the accuracy at 2mm from 72.4\% to 77.9\%, while also lowering ARE and increasing $\delta_1$. Furthermore, \figref{fig:qualitative_comparison} illustrates qualitative results on the StereoKP dataset, highlighting regions where the confidence-aware MDE model produces more stable and coherent depth estimates compared to the baseline. These results demonstrate that integrating confidence information into the training process leads to substantially more accurate and reliable depth predictions, particularly in areas with occlusions, specular reflections, or ambiguous textures. The improvements observed with 3D instrument keypoints further confirm that the confidence-aware model provides geometrically consistent and precise depth estimates for surgical tool, which are often challenging for conventional MDE methods. Overall, these findings validate that the proposed approach enhances the robustness and trustworthiness of depth estimation in realistic and noisy MIS environments.

\begin{table}[t]
\centering
\caption{Evaluation metrics for baseline and proposed methods across datasets. Bold values shows best results for a given metric and dataset.}
\begin{tabular}{l@{\hspace{2em}}|c @{\hspace{2.5em}} c|c @{\hspace{2.5em}} c}
\hline
\textbf{Dataset} & \multicolumn{2}{c}{\textbf{DAv1-B} \textbf{Baseline}}& \multicolumn{2}{c}{\textbf{DAv1-B} \textbf{CA (proposed)}}\\
 & ARE $\downarrow$ & $\delta_1$ $\uparrow$ & ARE $\downarrow$ & $\delta_1$ $\uparrow$ \\
\hline
StereoKP & 12.41 & 85.83 & \textbf{8.86} & \textbf{94.14} \\
MicroCT-SE & 5.69 & \textbf{99.62} & \textbf{5.04} & 99.60 \\
MicroCT-PK & 18.64 & 96.98 & \textbf{14.74} & \textbf{97.27} \\
Hamlyn \cite{recasens2021endo} & 13.97 & 80.00 & \textbf{13.74} & \textbf{80.10} \\
DaVinci \cite{ye2017self} & 8.28 & 96.60 & \textbf{8.03} & \textbf{96.80} \\
\hline
\end{tabular}
\label{tab:main_experimental_comparison}
\end{table}

\section{Conclusion}
We presented a confidence-aware monocular depth estimation framework tailored for minimally invasive surgery. By leveraging ensemble-based stereo matching for pixel-wise confidence estimation and integrating confidence-aware loss functions and prediction heads, our approach enhances both the reliability and accuracy of depth predictions. Extensive experiments across internal and public datasets demonstrate consistent improvements in keypoint-based and dense depth metrics, with the most significant gains observed on the challenging, artifact-prone StereoKP dataset. Results on MicroCT data confirm close alignment with gold-standard measurements under controlled laboratory conditions, while incremental improvements on Hamlyn and DaVinci datasets highlight strong generalization across surgical domains. Overall, the proposed framework effectively addresses challenges of noise and uncertainty in surgical imagery, advancing the robustness and clinical applicability of MDE for computer-assisted interventions in minimally invasive surgery.

\section*{Statements and Declarations}
\begin{itemize}
\item Competing Interests: MA, EC, PM, NT, RS, MR, FB, RM, IL and DS are employees of Medtronic Digital Technologies. 
\item Ethical approval: Medtronic plc maintains all necessary rights and consents to
process, analyse, and display the private data referenced in this study.
\end{itemize}

\bibliography{sn-bibliography}
\end{document}